\documentclass[runningheads]{llncs}

\usepackage{accv}

\usepackage{accvabbrv}
\usepackage{graphicx}
\usepackage{booktabs}
\usepackage{amsmath}
\usepackage[breaklinks,colorlinks,citecolor=accvblue]{hyperref}

\begin{document}

\title{Interpretable Uncertainty Routing Separating Emotion Ambiguity from Distribution Shift in Facial Expression Recognition}

\titlerunning{Interpretable Uncertainty Routing for FER}

\author{Keito Inoshita$^{*}$\inst{1,2} \and Takato Ueno\inst{3}}
\authorrunning{K. Inoshita and T. Ueno}
\institute{Faculty of Business and Commerce, Kansai University, Suita, Japan \and
Data Science and AI Innovation Research Promotion Center, Shiga University, Hikone, Japan \and
Graduate School of Data Science, Shiga University, Hikone, Japan}

\maketitle

\begin{abstract}
Facial expression recognition (FER) is inherently ambiguous: human annotators frequently disagree, and models deployed in real environments face distribution shift. Crucially, these two conditions demand different downstream actions, as ambiguous in-distribution faces should be reported with their ambiguity whereas out-of-distribution inputs should be rejected. However, a single uncertainty score conflates the two. In this study, uncertainty decomposition into aleatoric and epistemic components for FER is investigated, and Uncertainty-Aware Routing (UAR), an inference-time routing mechanism that exploits the separation, is introduced. Specifically, aleatoric and epistemic uncertainties are obtained from a Deep Ensemble of fully fine-tuned DINOv2 models and are each validated against an independent external signal: aleatoric against human annotator disagreement, and epistemic against distribution shift induced by image corruptions. The proposed dual-validation protocol reveals that aleatoric recovers annotator disagreement with Spearman correlation 0.66 (95\% CI: 0.64--0.68), and epistemic detects corruption-induced shifts, achieving average AUROC of 0.699 at the highest corruption severity. UAR retains approximately 1.8 times more ambiguous in-distribution faces than single-uncertainty routing at a matched out-of-distribution rejection rate. A strong label-distribution-learning baseline achieves comparable disagreement recovery but cannot separate ambiguity from shift and therefore cannot route, establishing that the value of decomposition lies in the separation enabling interpretable and differentiated action selection.

\keywords{interpretable AI \and uncertainty decomposition \and facial expression recognition \and deep ensembles \and distribution shift}
\end{abstract}

\section{Introduction}
\label{sec:intro}

With the rapid development of deep learning, facial expression recognition (FER) has been widely adopted in human--computer interaction, driver monitoring, and affective computing. Unlike many recognition tasks, its target is subjective perception that varies across observers, and large-scale FER datasets are therefore annotated by multiple workers who provide label distributions rather than definitive ground-truth labels. FERPlus~\cite{barsoum2016ferplus} supplies ten votes per image across eight emotion categories; the resulting empirical voting distribution directly describes how much humans disagree about each face, and this disagreement reflects ambiguity that is intrinsic to the expression itself rather than annotation noise that should be suppressed~\cite{plank2022problem}.

This intrinsic ambiguity reveals two qualitatively distinct sources of uncertainty in FER. The first is aleatoric uncertainty, which corresponds to the fundamental ambiguity of the expression, is reflected in annotator disagreement, and cannot be reduced by collecting additional data. The second is epistemic uncertainty, which arises from insufficient model knowledge and is expected to increase as inputs depart from the training distribution. Bayesian deep learning provides a principled framework for decomposing these two components; however, recent research has shown that the two estimates can be entangled and that the semantic validity of each decomposed component depends not only on the decomposition formula itself but also on the adequacy of the estimation method and its approximation~\cite{depeweg2018decomposition}.

This challenge is particularly consequential for FER. Accuracy and calibration alone cannot distinguish whether uncertainty originates from an intrinsically ambiguous expression or from an unknown input. A single confidence score, even after temperature calibration~\cite{guo2017calibration}, conflates the two: low confidence may mean that humans also disagree about the expression, or it may mean that the model is operating out of distribution. This distinction is precisely what downstream systems require, as ambiguous but in-distribution faces should be reported with their ambiguity, whereas out-of-distribution inputs should be deferred to human review or rejected. Prior work has proposed decomposing aleatoric and epistemic uncertainty in subjective natural language processing tasks through cyclical SG-MCMC and soft-label learning~\cite{inoshita_mcmc}; the present study transfers and develops this decomposition perspective to FER using Deep Ensembles. Consequently, the central challenge is not to decompose uncertainty for its own sake but to assign externally verifiable meaning to each component and to convert that separation into interpretable and actionable presentation.

In this study, a framework for dual-validated uncertainty decomposition in facial expression recognition is designed and introduced, centred on a novel inference-time mechanism termed Uncertainty-Aware Routing (UAR) and its learned variant, Learned UAR (L-UAR). Specifically, aleatoric and epistemic uncertainties are obtained from a Deep Ensemble of fully fine-tuned DINOv2~\cite{oquab2024dinov2} models following the information-theoretic decomposition of Depeweg \etal~\cite{depeweg2018decomposition}, consistent with the framework of Kendall and Gal~\cite{kendall2017uncertainties}, and each component is validated against an independent external signal before being used for routing. The main contributions of this study are summarised as follows.
\begin{enumerate}
  \item[i)] UAR is newly designed, in which epistemic uncertainty rejects out-of-distribution (OOD) inputs and aleatoric uncertainty retains ambiguous in-distribution faces, achieving the joint objective of OOD rejection and ambiguity retention that is structurally impossible with any single uncertainty scalar, and its learned variant L-UAR is further constructed from decomposed uncertainty features and shown to generalise to unseen corruption types, outperforming both the fixed-threshold UAR and single-confidence routing.
  \item[ii)] A dual-validation protocol is developed in which each uncertainty component is grounded in an independent external signal, with aleatoric validated against human annotator disagreement and epistemic validated against distribution shift induced by image corruptions and cross-dataset transfer, and a strong label distribution learning (LDL) baseline~\cite{geng2016ldl} is shown to achieve comparable disagreement recovery while being unable to detect shift or perform routing, thereby clarifying that the value of decomposition lies in the separation it enables.
  \item[iii)] A framework for presenting decomposed uncertainty as interpretable human-actionable decision-making is developed by partitioning the $(H_{\text{ale}}, H_{\text{epi}})$ space into Accept, Defer, and Reject regions so that the reason for uncertainty (emotion ambiguity or distribution shift) is made explicit in a form that humans can interpret and act upon, and this framework is evaluated honestly in broad experiments with 95\% confidence intervals and significance tests.
\end{enumerate}
The rest of this paper is organised as follows. Section~\ref{sec:related} reviews related work. Section~\ref{sec:method} introduces the proposed methodology. Section~\ref{sec:experiments} presents experimental results. Section~\ref{sec:discussion} discusses key findings and limitations. Section~\ref{sec:conclusion} concludes the paper.

\section{Related Work}
\label{sec:related}

\subsection{Facial Expression Recognition under Label Ambiguity}

\begin{sloppypar}
Robust FER under noisy or ambiguous labels has been actively investigated. One prominent line suppresses uncertain labels. Self-Cure Network (SCN)~\cite{wang2020scn} reweights and relabels uncertain samples using self-attention importance scores and rank regularisation. Latent distribution mining~\cite{she2021dmue} recovers instance-level label distributions from multiple auxiliary branches, and relative uncertainty learning~\cite{zhang2021rul} handles sample difficulty relatively. Erasing Attention Consistency (EAC)~\cite{zhang2022eac} mitigates noise memorisation through random erasing and feature consistency regularisation. Recent advances include landmark-aware methods~\cite{wu2023lanet} and temporal distribution distillation~\cite{zhang2024mart}.
\end{sloppypar}

A second line embraces annotator disagreement through LDL. Rather than training on majority-vote labels, LDL methods directly minimise a divergence between the predicted distribution and the empirical voting distribution~\cite{geng2016ldl}, recovering richer supervision and improving generalisation~\cite{le2023uncertainty}. Methods that explicitly represent uncertainty include uncertainty-aware representation learning~\cite{zhou2025uafer} and direct handling of label ambiguity~\cite{lee2025nla}; strong visual feature backbones have further advanced FER accuracy~\cite{mao2025posterpp}. Prior work has also shown that LLM-generated emotion labels fail to reproduce human uncertainty distributions~\cite{inoshita2026}, motivating anchoring uncertainty to human annotations; structure in emotion transitions has additionally been discovered from multi-annotator disagreement~\cite{inoshita_bsetd}.

Both lines improve robustness or uncertainty recovery but share a common limitation: they supply a single per-instance uncertainty scalar that conflates ambiguity with distributional shift, so that an intrinsically ambiguous face and an OOD input both receive high uncertainty and are treated identically. The present study implements LDL as a strong baseline and demonstrates that it matches our decomposition on disagreement recovery while being unable to separate ambiguity from shift and therefore unable to route.

\subsection{Uncertainty Decomposition and Its Critique}

Aleatoric--epistemic decomposition is a cornerstone of Bayesian deep learning~\cite{kendall2017uncertainties,hullermeier2021aleatoric}. A method for decomposing aleatoric and epistemic uncertainty in subjective natural language processing tasks through cyclical SG-MCMC and soft-label learning has been proposed~\cite{inoshita_mcmc}; the present study transfers and develops this perspective to FER (images). Heteroscedastic formulations separate input-dependent noise from model uncertainty via learned variance heads~\cite{depeweg2018decomposition}, information-theoretic decomposition expresses epistemic uncertainty as mutual information~\cite{depeweg2018decomposition}, and Deep Ensembles provide strong predictive uncertainty~\cite{lakshminarayanan2017ensembles}. Under distribution shift, ensemble models are more robust than single-model softmax confidence~\cite{ovadia2019can}, and MC-Dropout offers a cheaper approximation~\cite{gal2016dropout}.

However, recent studies have questioned whether truly separated decomposition is achievable in practice~\cite{depeweg2018decomposition}: both components increase under corruptions, so aleatoric remaining constant under shift does not hold in absolute terms. The practical consequence is that decomposition cannot be validated by theory alone but must be evaluated against external signals. Broader concerns about Affective AI, from data collection to social impact, have also been raised~\cite{inoshita_silos}, underscoring the importance of validating decomposition meaning in its social context.

The present study responds to this critique by validating aleatoric and epistemic components each against an independent external signal: human annotator disagreement and distribution shift, respectively. This conditional separation, in which each component correlates preferentially with a distinct external phenomenon, constitutes a practical and falsifiable notion of meaning that does not require perfect orthogonality.

\subsection{Calibration, Human Disagreement, and Selective Prediction}

Temperature calibration aligns confidence with accuracy and is widely applied to post-hoc improve a model's reliability~\cite{guo2017calibration}. The maximum softmax probability serves as a standard baseline for misclassification and OOD detection~\cite{hendrycks2017baseline}. Benchmarking under image corruptions~\cite{hendrycks2019benchmarking} has shown that standard networks are poorly calibrated under shift, while ensemble methods and temperature scaling partially mitigate this~\cite{ovadia2019can}.

Selective prediction allows a model to abstain on uncertain inputs, reducing risk at the cost of coverage~\cite{geifman2017selective}. Confidence estimation based on failure prediction~\cite{corbiere2019failure} improves abstention criteria, and extensions to learning-to-defer~\cite{mao2023twostage} enable routing to multiple specialists. In subjective tasks such as emotion recognition, annotator disagreement is a genuine signal that should be preserved rather than suppressed~\cite{plank2022problem,uma2021learning}.

The present study builds on these findings by combining selective prediction with uncertainty decomposition, distinguishing the general rejection objective (where maximum probability is competitive) from the routing objective (where a single score is structurally insufficient), and reporting both settings honestly.

\section{Methodology}
\label{sec:method}

\begin{figure}[t]
  \centering
  \includegraphics[width=0.85\linewidth]{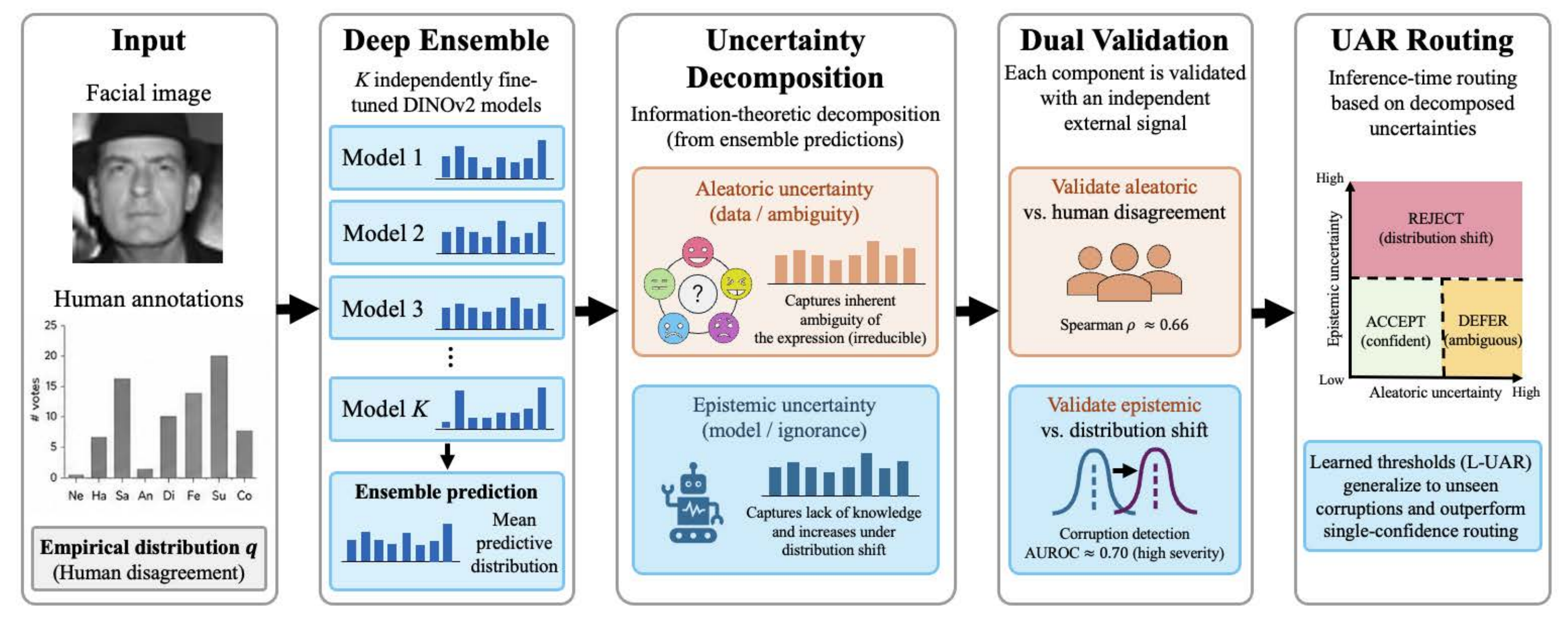}
  \caption{Overall pipeline of the proposed dual-validated uncertainty decomposition
           and routing framework.}
  \label{fig:overview}
\end{figure}

Figure~\ref{fig:overview} illustrates the overall pipeline of the proposed framework, which consists of a fully fine-tuned DINOv2 ensemble, information-theoretic decomposition, dual validation, and inference-time routing.

\subsection{Problem Definition and Notation}

Each sample consists of a facial image $x$ with $C$ emotion categories and an empirical distribution $q \in \Delta^{C}$ formed by the votes of $A$ annotators. The majority-vote label is $y = \arg\max_{c}\, q_{c}$, and human disagreement is characterised by $d = 1 - \max_{c} q_{c}$ and the entropy $H(q) = -\sum_{c} q_{c} \log q_{c}$. In FERPlus~\cite{barsoum2016ferplus}, $A = 10$ votes are available per image, providing an external description of aleatoric ambiguity that is independent of the model. The goal is to predict $y$ while obtaining an aleatoric estimate that tracks $d$ and an epistemic estimate that responds to distribution shift, so that the separation between the two components can be converted into routing actions at inference time.

\subsection{Deep Ensemble of Fully Fine-Tuned DINOv2 Models}

Meaningful epistemic uncertainty requires diversity across models, which cannot be obtained from a posterior distribution over a frozen head. Specifically, a frozen-head cyclical SG-MCMC approximation was tested and found to yield degenerate epistemic estimates that are insensitive to shift. Accordingly, $K$ self-supervised DINOv2 ViT-B/14 models~\cite{oquab2024dinov2,dosovitskiy2021vit} are independently fine-tuned end-to-end with different random seeds and data orderings. At inference time, member $k$ produces a predictive distribution $p_{k}(y \mid x)$, and the ensemble predictive distribution is the member average:
\begin{equation}
  \bar{p}(y \mid x) = \frac{1}{K} \sum_{k=1}^{K} p_{k}(y \mid x).
  \label{eq:ensemble_pred}
\end{equation}
In all experiments, $K = 5$. Full fine-tuning is essential: frozen ImageNet features are insufficient for separation (linear-probe accuracy falls below the majority-vote baseline), whereas fully fine-tuned DINOv2 achieves approximately 0.85 test accuracy on FERPlus. The five individual members attain test accuracies of 0.851, 0.825, 0.838, 0.776, and 0.853; the ensemble accuracy of 0.857 exceeds all individual members, confirming that even the weaker member contributes positively.

\subsection{Information-Theoretic Uncertainty Decomposition}

Following the information-theoretic decomposition of Depeweg \etal~\cite{depeweg2018decomposition}, consistent with the framework of Kendall and Gal~\cite{kendall2017uncertainties}, total predictive entropy is decomposed as:
\begin{align}
  H_{\text{tot}} &= H\!\left[\bar{p}(y \mid x)\right], \label{eq:H_tot} \\
  H_{\text{ale}} &= \frac{1}{K}\sum_{k=1}^{K} H\!\left[p_{k}(y \mid x)\right], \label{eq:H_ale} \\
  H_{\text{epi}} &= H_{\text{tot}} - H_{\text{ale}}. \label{eq:H_epi}
\end{align}
Here $H_{\text{ale}}$ is the mean within-member entropy, capturing intrinsic prediction ambiguity, while $H_{\text{epi}}$ is the mutual information between the ensemble index and the prediction, measuring member disagreement as a proxy for knowledge insufficiency. Fully end-to-end fine-tuned members produce non-degenerate $H_{\text{epi}}$ values that increase monotonically with corruption severity, in contrast to frozen-head variants.

\subsection{Dual-Validation Protocol}

Each component is validated against an independent external signal to assign practical, falsifiable meaning. Aleatoric uncertainty is validated via an annotator disagreement detection (ADD) task. Images whose voting entropy $H(q)$ exceeds the top-$\tau$ percentile are treated as positive examples of high disagreement, and the area under the ROC curve (AUROC) of $H_{\text{ale}}$ for this binary classification task, together with the Spearman correlation between $H_{\text{ale}}$ and $d$, measures how faithfully aleatoric uncertainty tracks annotator disagreement. Epistemic uncertainty is validated via a shift-detection task. Distribution shift is induced by image corruptions at five severity levels across five types, and by cross-dataset transfer to RAF-DB~\cite{li2017rafdb}, and the AUROC of $H_{\text{epi}}$ for separating in-distribution from OOD inputs measures epistemic validity. Because temperature calibration is a monotone transformation that does not change rankings, a temperature-calibrated single model~\cite{guo2017calibration} constitutes a strong baseline against which epistemic detection is compared.

\subsection{Uncertainty-Aware Routing}

\begin{figure}[t]
  \centering
  \includegraphics[width=0.85\linewidth]{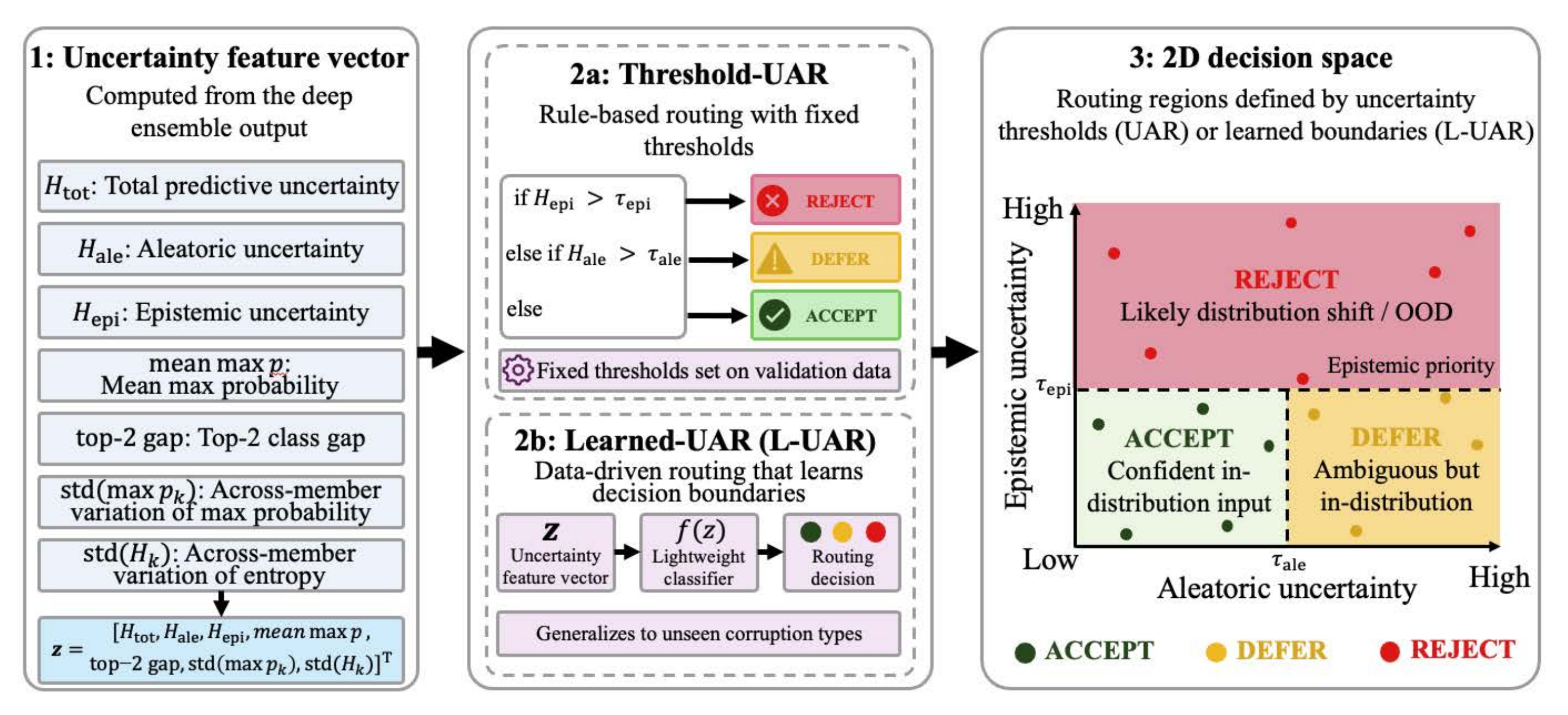}
  \caption{UAR routing mechanism: each input is assigned to one of three actions based on
           independent thresholds on $H_{\text{epi}}$ and $H_{\text{ale}}$.}
  \label{fig:uar}
\end{figure}

The separation between aleatoric and epistemic components is converted into an inference-time routing mechanism, as illustrated in Figure~\ref{fig:uar}.

For each input, the pair $(H_{\text{ale}}, H_{\text{epi}})$ is computed and the routing decision is made as follows: if $H_{\text{epi}} > \tau_{\text{epi}}$, the input is assigned to Reject (deferred to human review as out-of-distribution); otherwise, if $H_{\text{ale}} > \tau_{\text{ale}}$, the input is assigned to Defer (reported with its ambiguity as an ambiguous in-distribution face); otherwise the input is assigned to Accept (the argmax prediction is returned). Thresholds $\tau_{\text{epi}}$ and $\tau_{\text{ale}}$ are set on the validation split to match a target operating point, such as a desired OOD rejection rate.

It is noted that UAR achieves a routing objective that is structurally impossible with any single uncertainty scalar. For any monotone scalar $f(x)$, the high-$f$ set is the union of ambiguous in-distribution faces and OOD inputs, because both produce high predictive entropy regardless of the specific choice of $f$ (softmax entropy, maximum probability, temperature-scaled confidence, \etc). A single threshold on $f$ therefore inevitably rejects one type together with the other. By contrast, orthogonal thresholds in the two-dimensional space $(H_{\text{ale}}, H_{\text{epi}})$ structurally break this constraint, allowing high-epistemic inputs (OOD) and high-aleatoric, low-epistemic inputs (ambiguous in-distribution) to receive distinct actions. The routing rule is hierarchical, with epistemic taking priority: inputs that are simultaneously high in both $H_{\text{ale}}$ and $H_{\text{epi}}$ (ambiguous and OOD) are classified as Reject. This is a safety-side design that prevents OOD inputs from being incorrectly deferred to humans as ambiguous in-distribution faces, so that Accept and Defer exist only on the low-epistemic side ($H_{\text{epi}} \le \tau_{\text{epi}}$). The novelty of UAR lies not in the branching implementation but in enabling differentiated action selection that is impossible from a single score. Furthermore, the two-dimensional $(H_{\text{ale}}, H_{\text{epi}})$ space presents the Accept, Defer, and Reject regions in a visually and interpretably decomposable form, constituting an interpretable decision-presentation framework that allows humans to understand and act on the reason for uncertainty (emotion ambiguity or distribution shift).

To further improve routing, L-UAR is constructed by training a lightweight classifier on decomposed uncertainty features. For each sample, a feature vector is formed from total entropy $H_{\text{tot}}$, $H_{\text{ale}}$, $H_{\text{epi}}$, mean maximum probability, the top-two-class probability gap, across-member standard deviation of maximum probability, and across-member standard deviation of entropy. To demonstrate that L-UAR does not overfit to specific corruption types, the router is trained on a subset of corruption types and evaluated on held-out types not seen during training.

\section{Experiments}
\label{sec:experiments}

\subsection{Experimental Setup}

\textbf{Datasets.}
FERPlus~\cite{barsoum2016ferplus} provides ten votes per image across eight emotion categories and is used for training and aleatoric validation (test set: $n = 3{,}563$). RAF-DB~\cite{li2017rafdb} serves as a cross-dataset shift probe: its seven categories are mapped to the FERPlus label space and converted to the same greyscale format ($n = 3{,}068$). Eleven corruption types at five severity levels are applied to the FERPlus test set following Hendrycks and Dietterich~\cite{hendrycks2019benchmarking}: five core types (Gaussian noise, blur, centre occlusion, brightness, and pixelation) and six additional ImageNet-C types (shot noise, impulse noise, motion blur, contrast, JPEG compression, and elastic deformation).

\begin{sloppypar}
\textbf{Baselines.}
i) A single fine-tuned DINOv2 model. ii) A temperature-calibrated single model~\cite{guo2017calibration} using maximum softmax probability as confidence~\cite{hendrycks2017baseline}. iii) The Deep Ensemble ($K = 5$). iv) LDL~\cite{geng2016ldl}: DINOv2 fine-tuned by minimising KL divergence between predicted and empirical voting distributions. v) SCN~\cite{wang2020scn}: self-attention importance reweighting, rank regularisation, and relabelling, reproduced on our fine-tuned DINOv2 features. vi) EAC~\cite{zhang2022eac}: random erasing and flipped-feature consistency, reproduced on our fine-tuned DINOv2 features. SCN and EAC are included as representatives of standard FER uncertainty methods.
\end{sloppypar}

\textbf{Evaluation metrics.}
Accuracy, expected calibration error (ECE), Jensen--Shannon divergence to the human voting distribution, AUROC, and Spearman correlation are reported. All key values carry 95\% confidence intervals from 2,000 bootstrap iterations; differences in OOD detection are assessed with paired bootstrap tests. All experiments use a single NVIDIA H100 NVL GPU (PyTorch 2.2, timm 1.0, CUDA 12.1), with eight epochs, learning rate $3 \times 10^{-5}$, and batch size 128. Each ensemble member takes approximately 20 minutes to train. For reproducibility, all experiment scripts and configurations are released at \url{https://github.com/keito-git/interpretable-uncertainty-routing}.

\subsection{Analysis of Aleatoric Uncertainty and Annotator Disagreement}

\begin{figure}[t]
  \centering
  \includegraphics[width=0.7\linewidth]{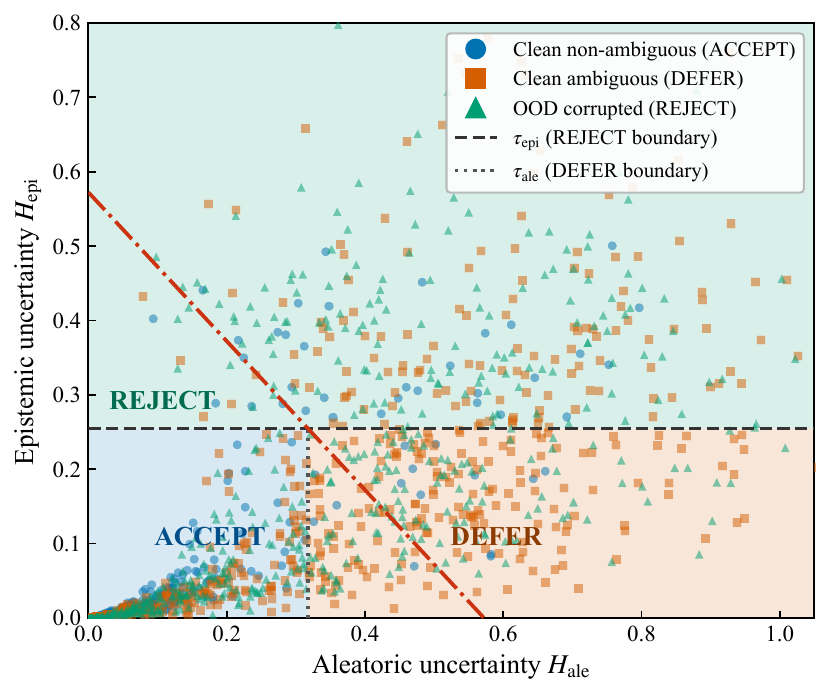}
  \caption{Scatter of decomposed uncertainties on clean FERPlus test images and OOD inputs,
           coloured by annotator disagreement; red dash-dot line: single-scalar threshold.}
  \label{fig:uar_concept}
\end{figure}

Figure~\ref{fig:uar_concept} shows the two-dimensional scatter of $(H_{\text{ale}}, H_{\text{epi}})$ on clean FERPlus test images and OOD inputs. The ADD task evaluates whether an uncertainty score identifies high annotator disagreement as high-valued.

As shown in Figure~\ref{fig:uar_concept}, ambiguous faces with high annotator disagreement concentrate in the high-$H_{\text{ale}}$ region (Defer), while OOD inputs cluster in the high-$H_{\text{epi}}$ region (Reject). A single 1-D threshold groups Defer and Reject on the same side and cannot separate them; the two-dimensional decomposition structurally separates the two.

Table~\ref{tab:disagreement_recovery} quantitatively compares disagreement recovery. On the clean FERPlus test set, aleatoric correlates with human disagreement $d$ at Spearman $\rho = 0.657$ (95\% CI: $[0.637, 0.677]$), while epistemic reaches $\rho = 0.586$ ($[0.562, 0.609]$); non-overlapping intervals confirm aleatoric's advantage. ADD AUROC is 0.897 for aleatoric and 0.861 for epistemic, with the gap robust across positive-example thresholds from the top 20\% to 50\% of voting entropy (0.931 to 0.836). LDL achieves comparable recovery ($\rho = 0.671$, ADD AUROC 0.910); all ensemble members are trained with hard labels, so LDL holds a training-signal advantage and this comparison is conservative. SCN substantially underperforms ($\rho = 0.440$, ADD AUROC 0.804). Disagreement recovery is therefore achievable by single-scalar methods; the key limitation is that no single scalar can structurally separate ambiguous in-distribution faces from OOD inputs for routing.

\begin{table}[t]
  \centering
  \caption{Annotator disagreement recovery on the FERPlus test set.}
  \label{tab:disagreement_recovery}
  \begin{tabular}{lcc}
    \toprule
    Method & Spearman $\rho$ & ADD AUROC \\
    \midrule
    Ours (aleatoric)  & 0.657 {[}0.637, 0.677{]} & 0.897 \\
    Ours (epistemic)  & 0.586 {[}0.562, 0.609{]} & 0.861 \\
    LDL               & 0.671                     & 0.910 \\
    EAC               & 0.609                     & 0.878 \\
    SCN               & 0.440                     & 0.804 \\
    \bottomrule
  \end{tabular}
\end{table}

\subsection{Evaluation of Distribution Shift Detection}

\begin{figure}[t]
  \centering
  \includegraphics[width=\linewidth]{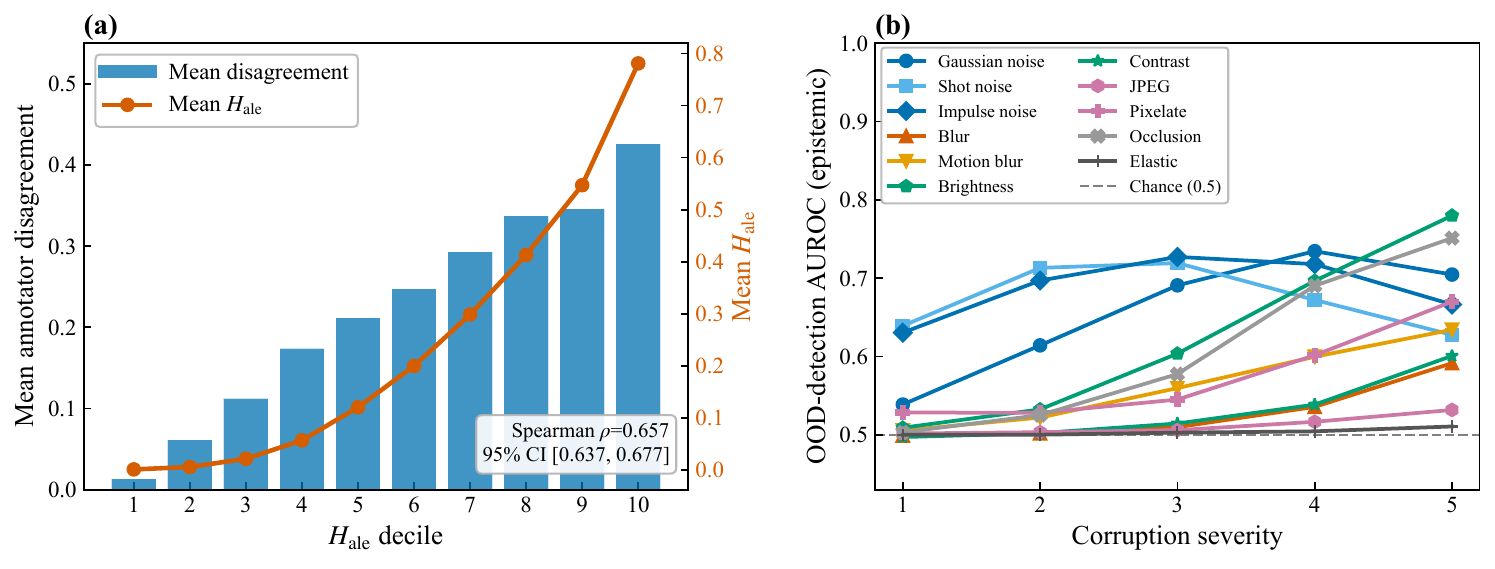}
  \caption{Dual-validation results: panel (a) $H_{\text{ale}}$ deciles vs.\ mean annotator
           disagreement; panel (b) OOD detection AUROC by corruption severity and type.}
  \label{fig:dual_validation}
\end{figure}

Figure~\ref{fig:dual_validation} summarises the dual-validation results. Panel (a) shows that $H_{\text{ale}}$ increases monotonically with annotator disagreement across deciles (Spearman $\rho = 0.657$), confirming that aleatoric accurately captures variability in annotator interpretation. Panel (b) shows OOD detection AUROC across corruption severities and types.

Under corruption, single models become overconfident: accuracy falls from 0.851 to 0.608 at the highest severity and ECE rises to 0.269. As shown in Figure~\ref{fig:dual_validation}(b), decomposed epistemic increases broadly with corruption severity, achieving average AUROC 0.699 at the highest severity across the five core corruption types, exceeding single-model confidence at 0.663. Because temperature calibration is monotone and preserves rankings, epistemic detection remains superior to the temperature-calibrated baseline. Paired bootstrap tests at the highest severity confirm significant advantages for Gaussian noise ($+0.077$; 95\% CI $[0.066, 0.089]$), blur ($+0.034$; $[0.024, 0.043]$), and pixelation ($+0.111$; $[0.101, 0.121]$). For brightness the advantage is $+0.002$ (non-significant), and for occlusion it is $-0.042$ ($[-0.053, -0.032]$): epistemic is inferior for occlusion. Across all eleven corruptions, decomposed epistemic exceeds single maximum probability for 9 out of 11 corruptions (mean AUROC 0.588 vs.\ 0.547), particularly strong for noise-type corruptions (impulse noise 0.688, Gaussian noise 0.657) and near chance for mild structural corruptions (elastic deformation 0.504, contrast 0.531).

Separation manifests in correlation direction rather than absolute magnitudes. Aleatoric wins on disagreement detection (AUROC 0.897 vs.\ 0.861) and epistemic wins on shift detection (mean AUROC 0.599 vs.\ aleatoric 0.587). The LDL baseline achieves only AUROC 0.526 on shift detection, essentially at chance, confirming that it cannot separate ambiguity from shift.

\subsection{Routing Evaluation}

\begin{figure}[t]
  \centering
  \includegraphics[width=\linewidth]{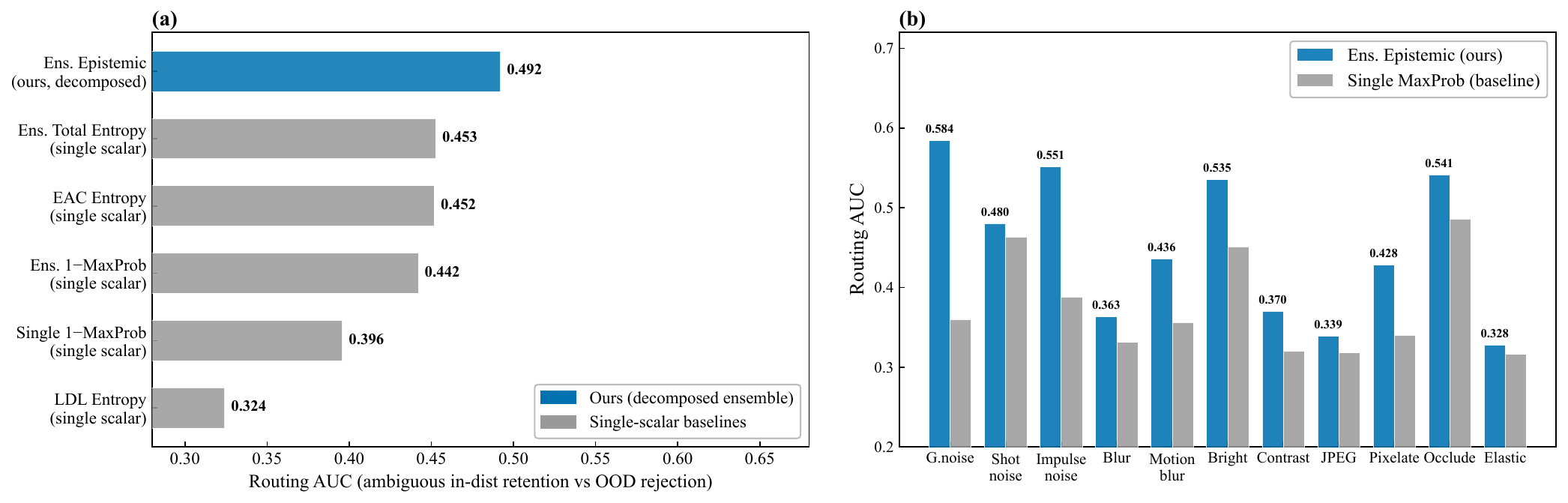}
  \caption{Routing performance comparison: panel (a) aggregate routing AUC across all
           methods; panel (b) per-corruption routing AUC for decomposed epistemic versus
           single maximum probability.}
  \label{fig:routing_compare}
\end{figure}

Figure~\ref{fig:routing_compare} presents the routing performance comparison. Panel (a) shows aggregate AUROC across all methods, and panel (b) shows decomposed epistemic versus single maximum probability across eleven individual corruptions.

UAR is evaluated on a pool mixing clean FERPlus images (in-distribution) with corrupted images (out-of-distribution) in equal proportion. At a matched OOD rejection rate of 0.70, epistemic-based routing retains 0.180 of ambiguous in-distribution faces versus 0.098 for single maximum probability, approximately 1.8 times higher. Ambiguous-retention versus OOD-rejection AUROC is 0.424 for epistemic vs.\ 0.376 for single maximum probability; the difference of 0.048 (95\% CI $[0.041, 0.071]$) is significant by paired bootstrap.

As shown in Figure~\ref{fig:routing_compare}(a), aggregate AUROC (30-seed average) shows decomposed epistemic at 0.492 outperforming ensemble predictive entropy 0.453, EAC entropy 0.452, single maximum probability 0.396, and LDL entropy 0.324. In the extended evaluation of Figure~\ref{fig:routing_compare}(b), decomposed epistemic exceeds single maximum probability for all eleven corruptions (11/11; aggregate AUROC 0.45 vs.\ 0.375).

\begin{sloppypar}
L-UAR further improves routing by learning from decomposed features. Trained on three corruption types (Gaussian noise, blur, occlusion) and evaluated on two unseen types (brightness, pixelation), L-UAR achieves AUROC 0.476 vs.\ 0.431 for fixed-threshold epistemic and 0.377 for single maximum probability. In leave-one-corruption-out evaluation, L-UAR consistently leads (averages: L-UAR 0.457, fixed-threshold 0.430, single maximum probability 0.374).
\end{sloppypar}

\subsection{Cross-Dataset Evaluation}

Table~\ref{tab:rafdb_ood} summarises cross-dataset and natural OOD evaluation on RAF-DB. Transfer accuracy is approximately 0.71 and the ensemble is better calibrated than the uncalibrated single model (ECE 0.146 vs.\ 0.218).

For truly OOD non-face images (176 images), epistemic effectively detects shift: mean epistemic rises from 0.104 to 0.232 and AUROC reaches 0.754. For 81 ``Unknown'' (ambiguous) images, aleatoric rises from 0.245 to 0.555, providing an additional positive finding that aleatoric captures natural annotation ambiguity.

On RAF-DB itself, however, epistemic does not fire as OOD (AUROC 0.538). This is interpreted as a desirable selectivity property rather than a failure. Maximum mean discrepancy (MMD) quantifies that linear MMD$^2$ between FERPlus and RAF-DB is 110.9, less than half of severity-4 Gaussian noise (231.3), showing that greyscale-aligned RAF-DB constitutes a benign domain transfer; epistemic responds to substantial shifts while not overreacting to benign transfer. For general rejection, single maximum probability (AUROC 0.887) is strong, consistent with the findings discussed in Section~\ref{sec:discussion}.

\begin{table}[t]
  \centering
  \caption{Cross-dataset and natural OOD evaluation results.}
  \label{tab:rafdb_ood}
  \begin{tabular}{lc}
    \toprule
    Metric & Value \\
    \midrule
    RAF-DB transfer accuracy           & 0.71 \\
    ECE (ensemble)                     & 0.146 \\
    ECE (single model)                 & 0.218 \\
    Epistemic OOD-AUROC (RAF-DB)       & 0.538 \\
    MMD$^2$ (linear, FER vs.\ RAF-DB)       & 110.9 \\
    MMD$^2$ (linear, FER vs.\ gnoise s4)    & 231.3 \\
    Epistemic AUROC (non-face, $n=176$)     & 0.754 \\
    \bottomrule
  \end{tabular}
\end{table}

\section{Discussion}
\label{sec:discussion}

\subsection{What Decomposition Adds over Single Uncertainty}

The central finding of this study is that disagreement recovery itself can be achieved by a single uncertainty score, notably by LDL, but decomposition provides an equally strong aleatoric component while simultaneously supplying a separable shift detector (epistemic). This separation enables UAR, realising the joint objective of OOD rejection and ambiguity retention that is structurally impossible from any single scalar. The interpretability offered by the partition of the $(H_{\text{ale}}, H_{\text{epi}})$ space into regions also presents the reason for uncertainty in a form that humans can interpret and act upon. This interpretability is the core value of the present study. The value of decomposition therefore lies not in accuracy or calibration but in enabling the action-differentiated, interpretable inference that characterises UAR.

Furthermore, the learned variant L-UAR demonstrates that decomposed features support generalisation to unseen corruption types. The router exploits multiple features simultaneously: maximum probability, inter-class gap, member variance, and epistemic uncertainty. As a result, L-UAR achieves higher retention of ambiguous in-distribution faces under novel shifts than either fixed-threshold UAR or single-confidence routing. This generalisation property is not achievable from a single scalar, since such a score cannot distinguish the reason for elevated uncertainty.

\subsection{Conditional Separation and the Scope of Single-Confidence Routing}

Both aleatoric and epistemic uncertainties increase in absolute magnitude under corruption, so complete orthogonal separation does not hold. This observation is consistent at the level of scale with the critique of information-theoretic decompositions raised by recent theoretical analyses~\cite{depeweg2018decomposition}. Nevertheless, conditional separation is achieved: aleatoric correlates preferentially with annotator disagreement and epistemic correlates preferentially with distribution shift. This constitutes a practical, falsifiable notion of meaning that does not require perfect orthogonality and provides an empirical answer to the question of whether decomposition is meaningful in this setting. It is noted that this conditional separation also rationalises the L-UAR results: the router succeeds because the decomposed features carry different information, not because they are perfectly orthogonal.

On the other hand, temperature-calibrated single-model confidence remains competitive for calibration quality under shift and actually outperforms the ensemble at the highest corruption severity (ECE 0.151 vs.\ 0.170). For general rejection (the standard selective prediction objective), maximum softmax probability yields a risk--coverage AUROC of 0.073, lower than the epistemic-based predictor at 0.086. Thus, for practitioners whose sole objective is to reduce prediction error through abstention, single-model confidence with temperature calibration is a strong and computationally cheaper alternative. Decomposition is beneficial specifically when the practitioner needs to distinguish between the two types of uncertain inputs and act differently on each.

\subsection{Limitations}
\label{sec:limitations}

Several limitations of the present study should be acknowledged. First, epistemic uncertainty responds to substantial corruptions and to natural non-face inputs, but does not detect benign cross-dataset variation (RAF-DB), and its OOD detection advantage is corruption-type dependent, being inferior to single maximum probability for occlusion. Second, aleatoric validation is limited to FERPlus, the only widely available FER dataset with per-image voting distributions. Third, SCN and EAC are reproduced on our fine-tuned DINOv2 features for stability rather than their original full-backbone implementations, though this is valid for the routing comparison. Fourth, the Deep Ensemble incurs five times the inference cost of a single model, although $K = 3$ is sufficient for most applications. Fifth, all experiments are conducted on low-resolution static images; extension to high-resolution images and video is left for future work. The value of decomposition lies in separation and routing: calibration and general rejection are regimes in which decomposition does not show an advantage.

\textbf{Ethics.}
FERPlus and RAF-DB are publicly released face image datasets used in accordance with their respective licences. Non-face images in FERPlus are defined by the annotators' majority vote and contain no identifiable individuals. No personally identifying information is collected or processed in this study.

\section{Conclusion}
\label{sec:conclusion}

This paper addresses the problem that a single confidence score cannot distinguish intrinsically ambiguous facial expressions from out-of-distribution inputs in facial expression recognition. To this end, a dual-validated uncertainty decomposition framework is introduced, in which aleatoric and epistemic uncertainties are obtained from a Deep Ensemble of fully fine-tuned DINOv2 models, validated against independent external signals, and converted into the inference-time Uncertainty-Aware Routing (UAR) mechanism and its learned variant (L-UAR). Experiments show that aleatoric uncertainty recovers annotator disagreement at Spearman $\rho = 0.66$, that decomposed epistemic uncertainty detects distribution shift at average AUROC 0.699, and that UAR retains approximately 1.8 times more ambiguous in-distribution faces than single-confidence routing at a matched rejection rate. A strong label-distribution-learning baseline achieves comparable disagreement recovery yet cannot separate ambiguity from shift or perform routing, establishing that the value of decomposition lies in the separation it enables.

These findings demonstrate the practical value of decomposition through interpretable uncertainty presentation in settings where emotion-AI systems must act differently on the two types of uncertainty. In future work, we will evaluate epistemic uncertainty under more realistic distribution shifts, extend UAR to high-resolution images and video sequences, and integrate the separated uncertainty components into human--AI collaborative systems.

\bibliographystyle{splncs04}
\bibliography{main}

\end{document}